\documentclass{article}

\usepackage[preprint,nonatbib]{neurips_2021}

\usepackage[utf8]{inputenc} 
\usepackage[T1]{fontenc}    
\usepackage{hyperref}       
\usepackage{url}            
\usepackage{booktabs}       
\usepackage{amsfonts}       
\usepackage{nicefrac}       
\usepackage{microtype}      
\usepackage{xcolor}         

\usepackage{bm}
\usepackage{amsmath}
\usepackage{amsthm}
\usepackage{color}
\usepackage{authblk}
\usepackage{multirow}
\usepackage{graphicx, float, url}
\usepackage{tikz}
\usepackage{breqn}
\usepackage{algorithm}
\usepackage{algorithmicx}
\usepackage[noend]{algpseudocode}
\usepackage{soul}
\usepackage{subfig}
\usepackage{threeparttable}

\newcommand{\cR}{\mathcal{R}}
\makeatletter
\def\blfootnote{\gdef\@thefnmark{}\@footnotetext}
\makeatother

\title{Privacy-Preserving Machine Learning with Fully Homomorphic Encryption for Deep Neural  Network}

  \author[1,*]{Joon-Woo Lee}
  \author[2,*]{HyungChul Kang}
  \author[2]{Yongwoo Lee}
  \author[2]{Woosuk Choi} 
  \author[2]{Jieun Eom}
  \author[2]{Maxim Deryabin} 
  \author[1]{Eunsang Lee }
  \author[1]{Junghyun Lee} 
  \author[2]{Donghoon Yoo }
  \author[3]{Young-Sik Kim} 
  \author[1]{Jong-Seon No} 
  \affil[1]{ 
    Dept. of Electrical and Computer Eng., INMC, Seoul National University, Korea, Republic of
    \protect\\
    \texttt{\{joonwoo3511, eslee3209, jhlee\}@ccl.snu.ac.kr},
    \texttt{jsno@snu.ac.kr}
  }
  \affil[2]{
    Samsung Advanced Institute of Technology, Korea, Republic of 
    \protect\\
    \texttt{\{hc1803.kang, yw0803.lee, woosuk0.choi, jieun.eom, max.deriabin, say.yoo\}@samsung.com}
  }
  \affil[3]{
    Dept. of Information and Communication Eng., Chosun University, Korea, Republic of 
    \protect\\
    \texttt{iamyskim@chosun.ac.kr}
  }

\begin{document}

\maketitle

\begin{abstract}
\blfootnote{*The first two authors contributed equally.}
Fully homomorphic encryption (FHE) is one of the prospective tools for privacy-preserving machine learning (PPML), and several PPML models have been proposed based on various FHE schemes and approaches. Although the FHE schemes are known as suitable tools to implement PPML models, previous PPML models on FHE encrypted data are limited to only simple and non-standard types of machine learning models. These non-standard machine learning models are not proven efficient and accurate with more practical and advanced datasets. Previous PPML schemes replace non-arithmetic activation functions with simple arithmetic functions instead of adopting approximation methods and do not use bootstrapping, which enables continuous homomorphic evaluations. Thus, they could not use standard activation functions and could not employ a large number of layers. The maximum classification accuracy of the existing PPML model with the FHE for the CIFAR-10 dataset was only 77\% until now. In this work, we firstly implement the standard ResNet-20 model with the RNS-CKKS FHE with bootstrapping and verify the implemented model with the CIFAR-10 dataset and the plaintext model parameters. Instead of replacing the non-arithmetic functions with the simple arithmetic function, we use state-of-the-art approximation methods to evaluate these non-arithmetic functions, such as the ReLU, with sufficient precision \cite{MinimaxRelu}. Further, for the first time, we use the bootstrapping technique of the RNS-CKKS scheme in the proposed model, which enables us to evaluate a deep learning model on the encrypted data. We numerically verify that the proposed model with the CIFAR-10 dataset shows 98.67\% identical results to the original ResNet-20 model with non-encrypted data. The classification accuracy of the proposed model is 90.67\%, which is pretty close to that of the original ResNet-20 CNN model. It takes about 4 hours for inference on a dual Intel Xeon Platinum 8280 CPU (112 cores) with 512 GB memory. We think that it opens the possibility of applying the FHE to the advanced deep PPML model.
\end{abstract}

\section{Introduction}
The privacy-preserving issue is one of the most practical problems for machine learning recently. Fully homomorphic encryption (FHE) is the most appropriate tool for privacy-preserving machine learning (PPML) to ensure strong security in the cryptographic sense and satisfy the communication's succinctness. FHE is an encryption scheme whose ciphertexts can be processed with any deep Boolean circuits or arithmetic circuits without access to the data. The security of FHE has been usually defined with indistinguishability under chosen-plaintext attack (IND-CPA) security, which is a standard cryptographic security definition. If the client sends the public keys and the encrypted data with an FHE scheme to the PPML server, the server can perform all computation needed in the desired service before sending the encrypted output to the client. Therefore, the application of FHE to PPML has been researched much until now.

The most successful PPML model on the homomorphically encrypted data until now was constructed with the TFHE homomorphic encryption scheme by Lou and Jiang \cite{SHE}, but it used the leveled version of the TFHE scheme without bootstrapping, which is not an FHE scheme. In other words, they chose in advance the parameters that can be used to perform the desired network without bootstrapping. If we want to design a deeper neural network with the leveled homomorphic encryption scheme, much impractically larger parameters have to be used, and it causes heavy run-time or memory overhead. Further, since the packing technique cannot be applied easily in the TFHE scheme, it can cause additional inefficiency with regard to the running time and the memory overhead if we want to process many data at once. Thus, it is desirable to use the FHE with moderate parameters and bootstrapping, which naturally supports the packing technique in the PPML model.

The applicable FHE schemes with this property are word-wise FHE schemes, such as Brakerski-Fan-Vercauteren (BFV) scheme \cite{bib_fv} or Cheon-Kim-Kim-Song (CKKS) scheme \cite{bib_ckks,CHK+18_rns}. Especially, the CKKS scheme has gained lots of interest for a suitable tool of the PPML implementation since it can deal with the encrypted real number naturally. However, these schemes support only homomorphic arithmetic operations such as the homomorphic addition and the homomorphic multiplication. Unfortunately, the popular activation functions are usually non-arithmetic functions, such as ReLU, sigmoid, leaky ReLU, and ELU. Thus, these activation functions cannot be evaluated directly using the word-wise FHE scheme. When the previous machine learning models using FHE replaced the non-arithmetic activation function with the simple polynomials, these models were not proven to show high accuracy for advanced classification tasks beyond the MNIST dataset.

Even though many machine learning models require multiple deep layers for high accuracy, there is no choice but to use a small number of layers in previous FHE-based deep learning models because FHE schemes' fast and accurate bootstrapping techniques are very recently available. The bootstrapping technique transforms a ciphertext that cannot support the homomorphic multiplication further to a fresh ciphertext by extending the levels of the ciphertext \cite{Genrty09,CHK+18_boot}. However, the bootstrapping technique has been actively improved in regard to algorithmic time complexity \cite{CCS19,bib_better,bib_non_sparse}, precision \cite{LLL+20}, and implementation \cite{GPUBootstrapping}, which make bootstrapping more practical. The PPML model with many layers has to be implemented with the precise and efficient bootstrapping technique in FHE.

\subsection{Our contribution}
In this paper, we firstly implement the ResNet-20 model for the CIFAR-10 dataset \cite{CIFAR10} using the residue number system CKKS (RNS-CKKS) \cite{CHK+18_rns} scheme, which is a variant of the CKKS scheme using the SEAL library 3.6.1 version \cite{SEAL21}, one of the most reliable libraries implementing the RNS-CKKS scheme. ResNet is one of the historic convolutional neural network (CNN) models which enables a very deep neural network with high accuracy for complex datasets such as CIFAR-10 and ImageNet. Many high-performance works for image classification are based on the ResNet model since these models can reach sufficiently high classification accuracy by stacking more layers. We firstly apply the ReLU function based on the composition of minimax approximate polynomials \cite{MinimaxRelu} to the encrypted data. Using the results, we firstly show the possibility of applying the FHE with the bootstrapping to the standard deep machine learning model by implementing ResNet-20 over the RNS-CKKS scheme. We use the RNS-CKKS bootstrapping with the SEAL library \cite{SEAL21}. The SEAL library is one of the most practical RNS-CKKS libraries, but it does not support the bootstrapping technique. The used bootstrapping can support sufficiently high precision to successfully use the bootstrapping in the ResNet-20 with the RNS-CKKS scheme for the CIFAR-10 dataset. 

Boemer et al. \cite{MP2ML} pointed out that all existing PPML models based on FHE or MPC are vulnerable to the model extraction attack. One of the reasons for this problem is that the previous PPML methods with the FHE scheme do not evaluate the softmax with the FHE scheme. It just sends the result before the softmax function, and then it is assumed that the client computes the softmax by itself. Thus, the information about the model can be extracted with lots of input-output pairs to the client. It can be desirable for the server to evaluate the softmax function with FHE. We firstly implement the softmax function in the machine learning model using the method in \cite{bib_ckks}, and this is the first implementation of a privacy-preserving machine learning model based on FHE preventing the model extraction attack.

We prepare the pre-trained model parameters by training the original ResNet-20 model with the CIFAR-10 plaintext dataset and perform the privacy-preserving ResNet-20 with these plaintext pre-trained model parameters and encrypted input images. We find that the inference result of the proposed implemented privacy-preserving ResNet-20 is 98.67\% identical to that of the original ResNet-20. It achieves 90.67\% classification accuracy, which is quite close to the original accuracy of 91.89\%. Thus, we verify that the proposed implemented PPML model successfully performs the ResNet-20 on the encrypted data. Further, the proposed implementation result shows the highest accuracy among the CNN implementation with the FHE scheme, while the previous highest result with the word-wise FHE scheme is only 77\% classification accuracy.

\subsection{Related works}
\paragraph{HE-friendly network}
Some previous works re-design the machine learning model compatible with the HE scheme by replacing the standard activation functions with the simple non-linear polynomials \cite{bib_cryptonet,bib_faster_cryptonet,SEALion,HCNN}, called the HE-friendly network. However, these machine learning models are usually successful only for the simple MNIST dataset and cannot reach sufficiently high accuracy for the CIFAR-10 dataset. The highest classification accuracy of the HE-friendly CNN with the simple polynomial activation function implemented by word-wise HE is only 77\% for the CIFAR-10 dataset \cite{HCNN}. Since the choice of the activation functions is sensitive in the advanced machine learning model, it may not be desirable to replace the standard and famous activation functions with simple arithmetic functions. Moreover, the additional pre-training process has to be followed before the PPML service is given. Since the training process is somewhat expensive in that it is pretty time-consuming and needs quite a lot of data, it is preferable to use the standard model such as ResNet and VGGNet for the plaintext data with its pre-trained model parameters when the data privacy has to be preserved.

\paragraph{Hybrid model with FHE and MPC}
Some previous works evaluate the non-arithmetic activation functions with the multiparty computation technique to implement the standard well-known machine learning model preserving privacy \cite{JVC18,Cheetah,ngraph1,ngraph2,MP2ML}. Although this method can evaluate even the non-arithmetic functions exactly, the privacy for the model information can be disclosed. In other words, the client should know the used activation function in the model, which is not desirable for the PPML servers. Also, since the communication with the clients is not succinct, the clients have to be involved in the computation, which is not desirable for the clients. 

\section{Preliminaries}
\subsection{RNS-CKKS scheme}
The CKKS scheme \cite{bib_ckks} is an FHE scheme supporting the arithmetic operations on encrypted data over real or complex numbers. Any users with the public key can process the encrypted real or complex data with the CKKS scheme without knowing any private information. The security of the CKKS scheme is based on the Ring-LWE hardness assumption. The supported homomorphic operations are the addition, the multiplication, the rotation, and the complex conjugation operation, and each operation except the homomorphic rotation operation is applied component-wisely. The rotation operation homomorphically performs a cyclic shift of the vector by some step. The multiplication, rotation, and complex conjugation operations in the CKKS scheme need additional corresponding evaluation keys and the key-switching procedures. Each real number data is scaled with some big integer, called the scaling factor, and then rounded to the integer before encrypting the data. When the two data encrypted with the CKKS scheme are multiplied homomorphically, the scaling factors of the two data are also multiplied. This scaling factor should be reduced to the original value using the rescaling operation for the following operations.

Since the CKKS scheme needs pretty big integers, the original CKKS scheme uses a multi-precision library, which requires higher computational complexity. To reduce the complexity, the residue number system variant of the CKKS scheme \cite{CHK+18_rns}, called the RNS-CKKS scheme, was also proposed. In the residue number system, the big integer is split into several small integers, and the addition and the multiplication of the original big integers are equivalent to the corresponding component-wise operations of the small integers. We use the RNS-CKKS scheme in this paper. 

\subsection{Bootstrapping of CKKS scheme}
The rescaling operation reduces both the scaling factor and the ciphertext modulus, which is necessary for each homomorphic multiplication. After several consecutive multiplications, the ciphertext modulus cannot be reduced further at some point. The bootstrapping operation of the CKKS scheme \cite{CHK+18_boot} transforms the ciphertext with too small modulus into the fresh ciphertext with large modulus without changing the message. Therefore, any arithmetic circuits with large multiplicative depth can be obtained using the bootstrapping operation.

Since the bootstrapping operation of the CKKS scheme is the most time-consuming and erroneous operation among all homomorphic operations of the CKKS scheme, there have been many works improving the running time and the precision of the bootstrapping of the CKKS scheme \cite{CCS19,bib_better,bib_non_sparse,LLL+20,GPUBootstrapping}. With the development of the bootstrapping of the CKKS schemes, it can be used in practical applications. However, research on applying FHE with the state-of-the-art bootstrapping technique to the privacy-preserving deep neural network has not been done yet.

The bootstrapping of the CKKS scheme starts with raising the modulus of the ciphertext. Since the message polynomial is added with the product of some integer polynomial and the base modulus, the modular reduction of the coefficients of the message polynomial should be performed homomorphically. The ciphertext is transformed into the converted ciphertext that has the coefficients of the message polynomial as slots, called \textsc{CoeffToSlot}. Next, we perform the homomorphic modular reduction to the converted ciphertext, called \textsc{ModReduction}. Then, it is transformed reversely into the ciphertext with the modular-reduced slots in the \textsc{ModReduction} operation as the coefficients of the message polynomial, called \textsc{SlotToCoeff}.
\begin{table}[t]
\centering
\caption{The specification of ResNet-20 (CIFAR-10)} \label{resnet_spec}
\begin{tabular}{ll||cccccc} 
\multicolumn{2}{c||}{Layer}    & Input Size         & \#Inputs & Filter Size      & \#Filters & Output Size        & \#Outputs  \\ 
\hline
\multicolumn{2}{l||}{Conv1}             & $32 \times 32$ & 3                 & $3 \times 3$ & 16                 & $32 \times 32$ & 16                  \\ 
\hline
\multirow{3}{*}{Conv2} & 2-1       & $32 \times 32$ & 16                & $3 \times 3$ & 16                 & $32 \times 32$ & 16                  \\ 
                       & 2-2       & $32 \times 32$ & 16                & $3 \times 3$ & 16                 & $32 \times 32$ & 16                  \\ 
                       & 2-3       & $32 \times 32$ & 16                & $3 \times 3$ & 16                 & $32 \times 32$ & 16                  \\ 
\hline
\multirow{5}{*}{Conv3} & 3-1-1    & $32 \times 32$ & 16                & $3 \times 3$ & 32                 & $16 \times 16$ & 32                  \\ 
                       & 3-1-2    & $16 \times 16$ & 32                & $3 \times 3$ & 32                 & $16 \times 16$ & 32                  \\ 
                       & 3-1-s & $32 \times 32$ & 16                & $1 \times 1$ & 32                 & $16 \times 16$ & 32                  \\ 
                       & 3-2       & $16 \times 16$ & 32                & $3 \times 3$ & 32                 & $16 \times 16$ & 32                  \\ 
                       & 3-3       & $16 \times 16$ & 32                & $3 \times 3$ & 32                 & $16 \times 16$ & 32                  \\ 
\hline
\multirow{5}{*}{Conv4} & 4-1-1    & $16 \times 16$ & 32                & $3 \times 3$ & 64                 & $8 \times 8$   & 64                  \\ 
                       & 4-1-2    & $8 \times 8$   & 64                & $3 \times 3$ & 64                 & $8 \times 8$   & 64                  \\ 
                       & 4-1-s & $16 \times 16$ & 32                & $1 \times 1$ & 64                 & $8 \times 8$   & 64                  \\ 
                       & 4-2       & $8 \times 8$   & 64                & $3 \times 3$ & 64                 & $8 \times 8$   & 64                  \\ 
                       & 4-3       & $8 \times 8$   & 64                & $3 \times 3$ & 64                 & $8 \times 8$   & 64                  \\ 
\hline
\multicolumn{2}{c||}{Average Pooling}   & $8 \times 8$   & 64                & $8 \times 8$ & 64                 & -                               & 64                  \\ 
\hline
\multicolumn{2}{c||}{Fully Connected}   & $64 \times 1$  & 1                 & -                             & -                  & -                               & 10                  \\
\hline
\end{tabular}
\end{table}
\section{ResNet-20 on RNS-CKKS scheme}
\subsection{Structure}
\begin{figure}[h]
	\begin{center}
		\includegraphics[width=0.9\linewidth]{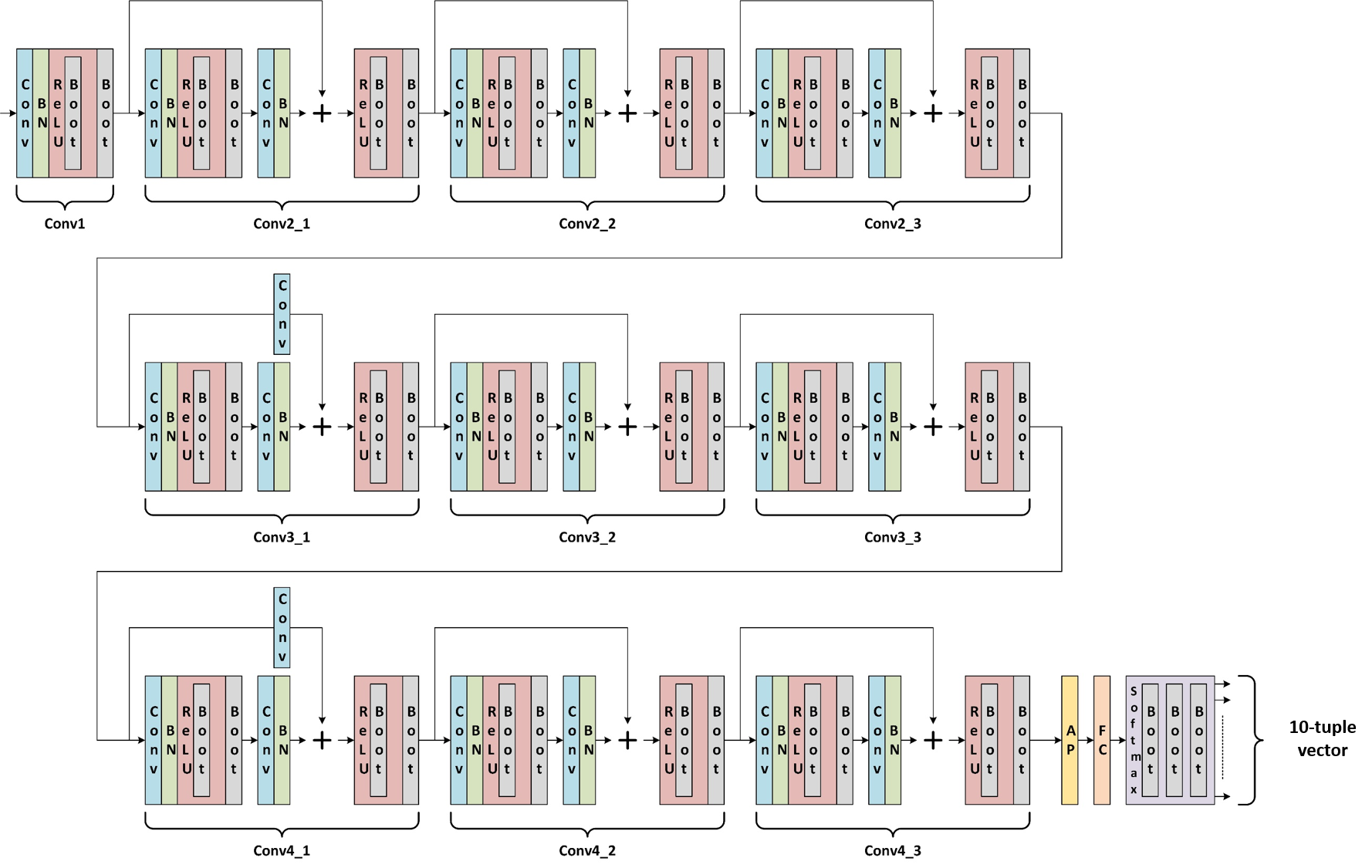}
		\caption{Structure of ResNet-20 over FHE.} \label{resnet_fhe}
	\end{center}
\end{figure}

We specify our implemented structure for the ResNet-20 with RNS-CKKS scheme as shown in Figure \ref{resnet_fhe}, where it consists of convolution (Conv), batch normalization (BN), ReLU, bootstrapping (Boot), average pooling (AP), fully connected layer (FC), and softmax. This model is virtually identical to the original ResNet-20 model except that the bootstrapping procedure is added. All of these procedures will be specified in the following subsections. Table \ref{resnet_spec} shows the specification of ResNet-20.

\subsection{General setting for RNS-CKKS scheme}
\subsubsection{Parameters}
We set the ciphertext polynomial degree as $2^{16}$ and the secret key Hamming weight as 64. The bit length of base modulus ($q_0$), special modulus, and default modulus are set as 60, 60, 50, respectively. The bit length of modulus in the bootstrapping range is the same as that of $q_0$. The numbers of levels for the general homomorphic operations and the bootstrapping are set as 11, 18, respectively. The maximum bit length of modulus is 1750, which satisfies 98-bit security. The security level $\lambda$ is computed based on Cheon et al.'s hybrid dual attack \cite{HybridDual}, which is the fastest attack for the LWE with the sparse key. Table \ref{seal_parameter} lists the above parameters. 

\begin{table}[htbp]
    \centering
    \caption{RNS-CKKS parameter settings} \label{seal_parameter}
    \begin{tabular}{cccccccccc}
        \multirow{2}{*}{$\lambda$} & Hamming & \multirow{2}{*}{Degree} & Modulus & \multirow{2}{*}{$q_{0}$} & Special & Scaling & Evaluation & Bootstrapping \\
        & Weight & & Q & & Prime & Factor & Level & Level \\
        \hline \hline
        98 & 64 & $2^{16}$ & 1750 bits & 60 bits & 60 bits & 50 bits & 11 & 18\\
        \hline
    \end{tabular}
\end{table}
\subsubsection{Data packing}
The message is a $32 \times 32$ CIFAR-10 image, and one image is processed at a time. We can use $2^{15}$ message slots in one ciphertext with our parameters, which is the half polynomial degree. Therefore, we employ the sparse packing method \cite{CHK+18_boot} to pack a channel of a CIFAR-10 image in one ciphertext using only $2^{10}$ sparse slots since the bootstrapping of sparsely packed ciphertext takes much less time than that of fully packed ciphertext.

\subsubsection{Data range and precision}
Any polynomials can approximate continuous functions only in some bounded set. If even one value in the message slots exceeds this bounded set, the absolute value of output diverges to a pretty big number, leading to complete classification failure. Since FHE can only handle arithmetic operations, polynomial approximation should be used for non-arithmetic operations such as the ReLU function, the bootstrapping, and the softmax function. Thus, the inputs for these procedures should be in the bounded approximation region. We analyze the absolute input values for the ReLU, the bootstrapping, and the softmax when performing the ResNet-20 with several images. Since the observed maximum absolute input value for these procedures is 37.1, we conjecture that the absolute input values for these procedures are less than 40 with a very high probability. We use this observation in the implementation of each procedure. We also empirically find that the precision of the approximate polynomial or the function should be at least $16$-bit below the decimal point, and thus we approximate each non-arithmetic function with 16-bit average precision.

\subsubsection{Optimization for precision of homomorphic operations}
We apply several methods to reduce the rescaling error and relinearization error and ensure the precision of the resultant message, such as the scaling factor management in \cite{KPP20}, lazy rescaling, and lazy relinearization \cite{BGP+20,LLK+20}. The lazy rescaling and relinearization can also be applied to reduce the computation time as it requires much computation due to the number-theoretic transformation (NTT) and gadget decomposition. 

\subsection{Convolution and batch normalization}

Most of the operations in the ResNet-20 are convolutions with zero-padded input to maintain their size.
We use the packed single input single output (SISO) convolution with stride 1 used in Gazelle \cite{JVC18}, which has low complexity for the encrypted data. Convolution with stride 2 is also required to perform down-sampling.
The striding convolution was also proposed by Gazelle \cite{JVC18} to decompose an input and a filter and perform convolutions and adding.

\begin{figure}[ht]
    \centering
    \begin{minipage}[b]{.45\textwidth}
        \subfloat[]{\includegraphics[width=0.55\linewidth]{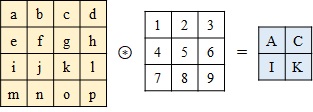}}
    \end{minipage}
    \begin{minipage}[b]{.45\textwidth}
        \subfloat[]{\includegraphics[width=\linewidth]{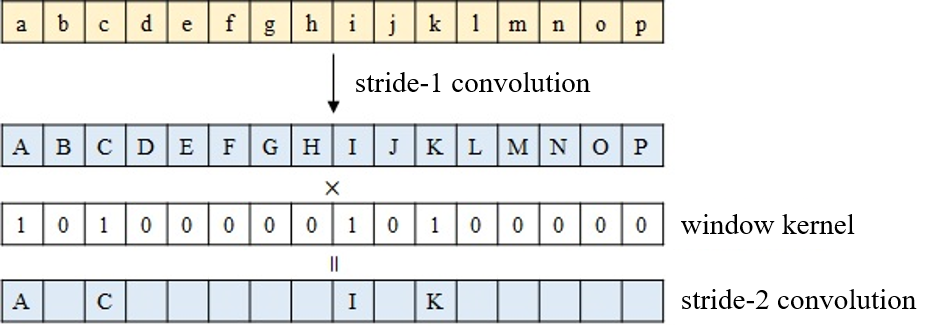}}
    \end{minipage}
    \caption{Stride-2 convolution (a) plaintext (b) ciphertext. }
    \label{ciphertext_convolution}
\end{figure}

We modify the striding convolution of Gazelle to reduce the required number of rotation operations. In the conventional striding convolution, we need to rearrange the encrypted data when decomposing the data, but the rearrangement increases the additional rotation operations. Since the rotation operation requires time-consuming key-switching procedures, it is desirable to reduce the number of rotation operations as much as possible. Instead of rearranging the slots to a well-structured channel, we perform the stride-2 convolution by extracting the valid values by multiplying the window kernel consisting of $0$ and $1$ from the stride-1 convolution result as shown in Figure \ref{ciphertext_convolution}:(b). It does not require any additional rotation operations. 

Since the batch normalization procedure is a simple linear function with constant coefficients, it can be implemented with the homomorphic addition and the homomorphic scalar multiplication.

\subsection{ReLU}
The activation function of the ResNet-20 is the ReLU function. Since the ReLU function is proven to be effective as the activation function in many CNN models, replacing the ReLU function with the simple arithmetic function such as $x^2$ \cite{bib_cryptonet} may not be desirable. Instead, we approximate the ReLU function by some arithmetic function with a sufficiently small error.

We firstly implement the ReLU function in the ResNet-20 with the RNS-CKKS scheme using the composition of the minimax polynomial approximation by Lee et al.\cite{MinimaxRelu}. To find an appropriate precision value, we repeatedly perform the ResNet-20 simulation over the RNS-CKKS scheme while changing the precision, and as a result, we find that the minimum 16-bit precision shows good performance on average. To synthesize the sign function for the ReLU approximation, we generate the composition of the small minimax approximate polynomials with precision parameter $\alpha = 13$ using the three minimax approximate polynomials with degrees 15, 15, and 27. This composition of polynomials ensures that the average approximation precision is about 16-bit precision. 

The homomorphic evaluation of the polynomials is carried out using the odd baby-giant method in \cite{HighPrecisionFirst} and the optimal level consumption method in \cite{bib_non_sparse}. Since the homomorphic evaluation of polynomial compositions consumes many depths, it is impossible to finish it without bootstrapping. Thus, we use bootstrapping twice, once in the middle and once at the end of evaluating the ReLU function. 

\subsection{Bootstrapping}
Since we have to consume many depths to implement the ResNet-20 on the RNS-CKKS scheme, many bootstrapping procedures are required to ensure enough homomorphic multiplications. For the first time, we apply the bootstrapping technique to perform the deep neural network such as the ResNet-20 on the encrypted data and prove that the FHE scheme with the state-of-the-art bootstrapping can be successfully applied for privacy-preserving deep neural networks. Since the \texttt{SEAL} library does not support any bootstrapping technique, we implement the most advanced bootstrapping with the \texttt{SEAL} library \cite{bib_non_sparse,LLL+20,GPUBootstrapping}. The \textsc{CoeffToSlot} and the \textsc{SlotToCoeff} are implemented using collapsed FFT structure \cite{CCS19} with depth 2. The \textsc{ModReduction} is implemented using the composition of the cosine function, two double angle formulas, and the inverse sine function \cite{bib_better,LLL+20}, where the cosine function and the inverse sine function are approximated with the multi-interval Remez algorithm as in \cite{LLL+20}. 

The most crucial issue when using the bootstrapping of the RNS-CKKS scheme is the bootstrapping failure. 1,149 bootstrapping procedures are required in our model, and the result of the whole neural network can be largely distorted if even one of the bootstrapping procedures fails. The bootstrapping failure occurs when one of the slots in the input ciphertext of the \textsc{ModReduction} procedure is not on the approximation region. The approximation interval can be controlled by the bootstrapping parameters $(K,\epsilon)$, where the approximation region is $\cup_{i=-(K-1)}^{K-1} [i-\epsilon,i+\epsilon]$ \cite{CHK+18_boot}. While the parameter $\epsilon$ is related to the range and the precision of the input message data, the parameter $K$ is related to the values composing the ciphertext and not related to the input data. Since the values contained in the ciphertext are not predictable, we have to investigate the relation between the bootstrapping failure probability and the parameter $K$.

\begin{table}[ht]
	\centering
	\caption{Boundary of approximation region given key Hamming weight and failure probability of modular reduction
		\label{table:failure_prob}
	}
	\begin{tabular}{c||ccc}
	     
		$\mathrm{Pr}(|I_i|\ge K)$ & $h=64$ & $h=128$ & $h=192$ \\\hline
		$2^{-23}$ \cite{bib_non_sparse}                 & $12$   & $17$    & $21$    \\
		$2^{-30}$                 & $14$   & $20$    & $24$    \\
		$2^{-40}$                 & $16$   & $23$    & $28$  \\
		\hline
	\end{tabular}
\end{table}

In Table \ref{table:failure_prob}, we show several bounds of the input message and its failure probability. The specific investigation is described in Appendix.
A larger bound means that a higher degree of the approximate polynomial is required; hence, more computation is required.
Using the new bound for approximation in Table \ref{table:failure_prob}, we can offer a trade-off between the evaluation time and failure probability of the whole network. Following \cite{bib_non_sparse,LLK+20}, the approximated modular reduction in the CKKS bootstrapping so far has failure probability $\approx 2^{-23}$, but it is not sufficiently small since we have to perform pretty many bootstrapping procedures for the ResNet-20. Thus, the bootstrapping failure probability is set to be less than $2^{-40}$ in our implementation. The Hamming weight of the secret key is set to be 64, and $(K,\epsilon)=(17,2^{-6})$. The corresponding degree for the minimax polynomial for the cosine function is 54, and that of the inverse sine function is 5, which is obtained by the multi-interval Remez algorithm \cite{LLL+20}.

\subsection{Average pooling and fully connected layer}
\begin{figure}[htbp]
	\begin{center}
		\includegraphics[width=\linewidth]{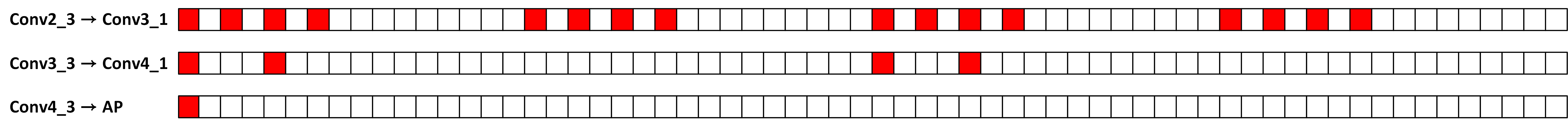}
		\caption{The valid message slots of the ciphertext.} \label{ap_fc}
	\end{center}
\end{figure}

After stride-$2$ convolution layers $3, 4$ and average pooling, the valid message slots of the ciphertext are given as follows (filled boxes in Figure \ref{ap_fc}).

If we combine these $64$ ciphertexts into one ciphertext using the rotation, we must perform the additional $63$ rotations when multiplying the matrix in the fully connected layer. Therefore, if we do not combine them into one, we can get an advantage in operation speed at the cost of more memory size. Thus, we do not need to perform the rotations when doing summation in the softmax.

\subsection{Softmax}
The last part of our implementation is the softmax function. Note that we have to evaluate the exponential function of each input value for the softmax function, and these output values can be too large to be embraced by the RNS-CKKS encryption scheme. Instead of evaluating $e^{x_i}/\sum_{j=1}^{10} e^{x_j}$ for each $i$, we evaluate $e^{x_i/4}/\sum_{j=1}^{10} e^{x_j/4}$. We find that it outputs almost the same output as the original softmax function, and the output of each exponential function does not exceed the capacity of the encryption scheme.

We approximate the exponential function $y = e^x$ by the polynomial with the degree of 12 by the least square method. The approximation region of the exponential function is set to be $[-1, 1]$. We multiply $x$ by a constant $1/64$ to put the input in the approximation region, evaluate the approximate polynomial of the exponential function, and square the output four times to evaluate $(e^{x/64})^{16}=e^{x/4}$. 

Then, the approximation method for the inverse function \cite{CKK+19} is applied to the sum of the exponential functions. The approximation region is $[0, 2]$, and the absolute value of inputs for the inverse is heuristically found to be less than $10^4$. Thus, we multiply the input by $10^{-4}$ before applying the approximate inverse function and multiply it again by $10^{-4}$ after the evaluation. Then, each exponential output is multiplied by this inverse output. These procedures for the softmax function need 22 bootstrapping procedures.

\section{Simulation result}
\subsection{Simulation setting and model parameters}
We simulate the proposed model by the SEAL library \cite{SEAL21} released by Microsoft. Our simulation environment is a dual Intel Xeon Platinum 8280 CPU (112 cores) with 512GB memory. We allocate one thread per one channel of each layer by using the OpenMP library to improve the execution speed of the ResNet-20.

The model parameters are prepared by the following training method. We use 32 x 32 color images, subtract the mean of the pixels in the training dataset, and adopt a data argumentation method such as shifting and mirroring horizontally for training. We adopt the He initialization \cite{HE} as the weight initialization and no dropout. We train the model with 32 $\times$ 32 mini-batches and cross-entropy loss function. The learning rate starts with a 0.001 learning rate divided by 10 after 80 epochs and 100 after 120 epochs during training. The classification accuracy with the trained model parameters is 91.89\%, which is tested with 10,000 images.

\subsection{Performance}
Table \ref{precision} shows the agreement ratio between the classification results of the implemented privacy-preserving ResNet-20 and that of the original ResNet-20, which shows almost the same results. We test the inference on 75 encrypted images, and the 95\% confidence interval is suggested for each result. Only one result of our proposed model for the encrypted data shows a different result from that of the ResNet-20 model for plaintext data. In other words, the agreement ratio is 98.67\%$\pm$ 2.59\%, which is a sufficiently high agreement result. The classification accuracy of the ResNet-20 for the encrypted data is 90.67\%$\pm$ 6.58\%, while that of the original ResNet-20 for the corresponding plaintext image is 89.33\%$\pm$ 6.99\%. Thus, we verify that the ResNet-20 can be successfully carried out using the RNS-CKKS scheme with sufficient accuracy for classification and the proper bootstrapping operation. Note that the highest classification accuracy of the previous model for the CIFAR-10 dataset over the word-wise FHE scheme is only 77\% \cite{HCNN}.

Table \ref{time} shows the running time for the whole ResNet-20 and the portion for each component in the model. The proposed model takes about 4 hours to infer one image, and the most time-consuming components are the convolution, the ReLU, and the bootstrapping.

\begin{table}[htbp]
  \centering
  \begin{threeparttable}
    \caption{Classification accuracy of ResNet-20 for plaintext and ciphertext and agreement ratio} \label{precision}
    \begin{tabular}{c||c||cc||c} 
    Model & ResNet-20\tnote{1} & ResNet-20\tnote{2} & PPML ResNet-20 & Agreement\\ 
    \hline
    Accuracy & 91.89\% $\pm$ 0.54\% & 89.33\% $\pm$ 6.99\% & 90.67\% $\pm$ 6.58\%& 98.67\%$\pm$ 2.59\% \\
    \hline
    \end{tabular}
    \begin{tablenotes}
      \small
      \item[1] Classification accuracy verified with 10,000 images.
      \item[2] Classification accuracy verified with 75 images which are used to test ResNet-20 on encrypted images.
    \end{tablenotes}
  \end{threeparttable}
\end{table}
\begin{table}[htbp]
    \centering
    \caption{The running time of ResNet-20 and the percentage of time spent in each component relative to total time} \label{time}
    \begin{tabular}{c||cccccc||c} 
    Layer & PC & BN & CR & Boot & AP + FC & Softmax & Total time (s)\\ 
    \hline
    Time ratio & 34.30\% & 0.20\% & 32.63\% & 29.97\% & 0.06\% & 2.83\% & 14,694\\
    \hline
    \end{tabular}
\end{table}
\section{Limitation}
\paragraph{Running time} The running time for the proposed model, which is about 4 hours, is somewhat large for practical use. This work firstly shows the possibility of applying the FHE to standard deep learning models with high accuracy, but it has to be optimized and improved in various ways to reduce the running time. Therefore, the essential future work is the advanced implementation of the ResNet-20 with the RNS-CKKS scheme with various accelerators realized using GPU, FPGA, or ASIC. Since research on implementing the state-of-the-art FHE scheme is advancing rapidly, the ResNet-20 over the encrypted data will be made more practically soon. Also, we implement the PPML model for only one image, and the running time per image can be much improved if we properly use the packing method of the RNS-CKKS scheme. We leave this optimization for many images as future work.

\paragraph{Security level} The security level of the proposed model is 98-bit security, which is the minimum security level that can be considered secure. Since the standard security level in most applications is 128 bit, someone can regard this security level as insecure, and we may want to raise the security level. However, this 98-bit security is not a hard limit of our implementation; we can easily raise the security level by changing the parameters of the RNS-CKKS scheme. This just makes the trade-off between the security and the running time, and thus we can reach the higher security level at the cost of longer running time.

\paragraph{Classification accuracy} Even if ML models are trained with the same hyper-parameters, the ML models have different performances because weights are initialized to random values for each training. Thus, the ML model performance, such as accuracy, is shown as the average values obtained by training several times. However, since we focus on implementing ResNet-20 for homomorphically encrypted data, we train this model only once, not many times. Nevertheless, we have shown that the encrypted ResNet-20 operation is possible with almost the same accuracy as the original ResNet-20 paper [Reference]. Furthermore, since it is implemented in the FHE with bootstrapping, it can be expected that the same result will be obtained for a deeper network than the Resnet-20.

\section{Conclusion}
For the first time, we applied the RNS-CKKS scheme, one of the state-of-the-art FHE schemes, to the standard deep neural network ResNet-20 to implement the PPML. Since the precise approximation of the ReLU function, the bootstrapping, and the softmax function had not been applied to the PPML models until now, we applied these techniques with fine-tuned various parameters. Then, we showed that the implemented ResNet-20 with the RNS-CKKS scheme achieves almost the same result as the original ResNet-20 and reaches the highest classification accuracy among the PPML models with the word-wise FHE scheme introduced so far. This work firstly suggested that the word-wise FHE with the most advanced techniques can be applied to the state-of-the-art machine learning model without re-training it.

{
\small
\bibliographystyle{unsrt}
\bibliography{main}

\begin{thebibliography}{10}

\bibitem{MinimaxRelu}
Junghyun Lee, Eunsang Lee, Joon-Woo Lee, Yongjune Kim, Young-Sik Kim, and
  Jong-Seon No.
\newblock Precise approximation of convolutional neural networks for
  homomorphically encrypted data.
\newblock arXiv preprint, abs/2105.10879, 2021.
\newblock \url{http://arxiv.org/abs/2105.10879}.

\bibitem{SHE}
Qian Lou and Lei Jiang.
\newblock {SHE}: A fast and accurate deep neural network for encrypted data.
\newblock {\em Advances in Neural Information Processing systems (NeurIPS)},
  2019.

\bibitem{bib_fv}
Junfeng Fan and Frederik Vercauteren.
\newblock Somewhat practical fully homomorphic encryption.
\newblock Cryptology ePrint Archive, Report 2012/144, 2020.
\newblock \url{https://eprint.iacr.org/2012/144}.

\bibitem{bib_ckks}
Jung~Hee Cheon, Andrey Kim, Miran Kim, and Yongsoo Song.
\newblock Homomorphic encryption for arithmetic of approximate numbers.
\newblock In {\em ASIACRYPT 2017}, pages 409--437, 2017.

\bibitem{CHK+18_rns}
Jung~Hee Cheon, Kyoohyung Han, Andrey Kim, Miran Kim, and Yongsoo Song.
\newblock A full {RNS} variant of approximate homomorphic encryption.
\newblock In {\em Proceedings of International Conference on Selected Areas in
  Cryptography (SAC)}, pages 347--368, Calgary, Canada, 2018.

\bibitem{Genrty09}
Craig Gentry.
\newblock Fully homomorphic encryption using ideal lattices.
\newblock In {\em Proceedings of the Forty-First Annual ACM Symposium on Theory
  of Computing (STOC)}, pages 169--178, 2009.

\bibitem{CHK+18_boot}
Jung~Hee Cheon, Kyoohyung Han, Andrey Kim, Miran Kim, and Yongsoo Song.
\newblock Bootstrapping for approximate homomorphic encryption.
\newblock In {\em EUROCRYPT 2018}, pages 360--384, 2018.

\bibitem{CCS19}
Hao Chen, Ilaria Chillotti, and Yongsoo Song.
\newblock Improved bootstrapping for approximate homomorphic encryption.
\newblock In {\em EUROCRYPT 2019}, pages 34--54, 2019.

\bibitem{bib_better}
Kyoohyung Han and Dohyeong Ki.
\newblock Better bootstrapping for approximate homomorphic encryption.
\newblock In {\em Proceedings of Cryptographers' Track at the RSA Conference},
  pages 364--390, 2020.

\bibitem{bib_non_sparse}
Jean-Philippe Bossuat, Christian Mouchet, Juan Troncoso-Pastoriza, and
  Jean-Pierre Hubaux.
\newblock Efficient bootstrapping for approximate homomorphic encryption with
  non-sparse keys.
\newblock Cryptology ePrint Archive, Report 2020/1203, 2020.
\newblock \url{https://eprint.iacr.org/2020/1203}, accepted to
  \textit{EUROCRYPT 2021}.

\bibitem{LLL+20}
Joon-Woo Lee, Eunsang Lee, Yongwoo Lee, Young-Sik Kim, and Jong-Seon No.
\newblock High-precision bootstrapping of {RNS-CKKS} homomorphic encryption
  using optimal minimax polynomial approxiamtion and inverse sine function.
\newblock Cryptology ePrint Archive, Report 2020/552, 2020.
\newblock \url{https://eprint.iacr.org/2020/552}, accepted to \textit{EUROCRYPT
  2021}.

\bibitem{GPUBootstrapping}
Wonkyung Jung, Sangpyo Kim, Jung~Ho Ahn, Jung~Hee Cheon, and Younho Lee.
\newblock Over 100x faster bootstrapping in fully homomorphic encryption
  through memory-centric optimization with gpus.
\newblock Cryptology ePrint Archive, Report 2021/508, 2021.
\newblock \url{https://eprint.iacr.org/2021/508}.

\bibitem{CIFAR10}
Alex Krizhevsky, Geoffrey Hinton, et~al.
\newblock Learning multiple layers of features from tiny images.
\newblock 2009.

\bibitem{SEAL21}
Microsoft.
\newblock Microsoft {SEAL}.
\newblock \url{https://github.com/microsoft/SEAL}, 2021.

\bibitem{MP2ML}
Fabian Boemer, Rosario Cammarota, Daniel Demmler, Thomas Schneider, and Hossein
  Yalame.
\newblock {MP2ML}: A mixed-protocol machine learning framework for private
  inference.
\newblock In {\em Proceedings of the 15th International Conference on
  Availability, Reliability and Security}, pages 1--10, 2020.

\bibitem{bib_cryptonet}
Ran Gilad-Bachrach, Nathan Dowlin, Kim Laine, Kristin Lauter, Michael Naehrig,
  and John Wernsing.
\newblock Cryptonets: Applying neural networks to encrypted data with high
  throughput and accuracy.
\newblock In {\em Proceedings of International Conference on Machine Learning
  (ICML)}, pages 201--210, 2016.

\bibitem{bib_faster_cryptonet}
Edward Chou, Josh Beal, Daniel Levy, Serena Yeung, Albert Haque, and
  Li~Fei-Fei.
\newblock Faster cryptonets: Leveraging sparsity for real-world encrypted
  inference.
\newblock arXiv preprint, abs/1811.09953, 2018.
\newblock \url{http://arxiv.org/abs/1811.09953}.

\bibitem{SEALion}
Tim van Elsloo, Giorgio Patrini, and Hamish Ivey-Law.
\newblock {SEALion}: A framework for neural network inference on encrypted
  data.
\newblock arXiv preprint, abs/1904.12840, 2019.
\newblock \url{http://arxiv.org/abs/1904.12840}.

\bibitem{HCNN}
Ahmad~Al Badawi, Jin Chao, Jie Lin, Chan~Fook Mun, Sim~Jun Jie, Benjamin
  Hong~Meng Tan, Xiao Nan, Aung Mi~Mi Khin, and Vijay Chandrasekhar.
\newblock Towards the {A}lexnet moment for homomorphic encryption: {HCNN}, the
  first homomorphic {CNN} on encrypted data with {GPU}s.
\newblock {\em IEEE Transactions on Emerging Topics in Computing}, 2020.

\bibitem{JVC18}
Chiraag Juvekar, Vinod Vaikuntanathan, and Anantha Chandrakasan.
\newblock {GAZELLE}: A low latency framework for secure neural network
  inference.
\newblock In {\em 27th USENIX Security Symposium}, pages 1651--1669, 2018.

\bibitem{Cheetah}
Brandon Reagen, Woo-Seok Choi, Yeongil Ko, Vincent~T Lee, Hsien-Hsin~S Lee,
  Gu-Yeon Wei, and David Brooks.
\newblock Cheetah: Optimizing and accelerating homomorphic encryption for
  private inference.
\newblock In {\em 2021 IEEE International Symposium on High-Performance
  Computer Architecture (HPCA)}, pages 26--39. IEEE, 2020.

\bibitem{ngraph1}
Fabian Boemer, Yixing Lao, Rosario Cammarota, and Casimir Wierzynski.
\newblock n{G}raph-{HE}: A graph compiler for deep learning on homomorphically
  encrypted data.
\newblock In {\em Proceedings of the 16th ACM International Conference on
  Computing Frontiers}, pages 3--13, 2019.

\bibitem{ngraph2}
Fabian Boemer, Anamaria Costache, Rosario Cammarota, and Casimir Wierzynski.
\newblock n{G}raph-{HE}2: A high-throughput framework for neural network
  inference on encrypted data.
\newblock In {\em Proceedings of the 7th ACM Workshop on Encrypted Computing \&
  Applied Homomorphic Cryptography}, pages 45--56, 2019.

\bibitem{HybridDual}
Jung~Hee Cheon, Minki Hhan, Seungwan Hong, and Yongha Son.
\newblock A hybrid of dual and meet-in-the-middle attack on sparse and ternary
  secret lwe.
\newblock {\em IEEE Access}, 7:89497--89506, 2019.

\bibitem{KPP20}
Andrey Kim, Antonis Papadimitriou, and Yuriy Polyakov.
\newblock Approximate homomorphic encryption with reduced approximation error.
\newblock Cryptology ePrint Archive, Report 2020/1118, 2020.
\newblock \url{https://eprint.iacr.org/2020/1118}.

\bibitem{BGP+20}
Marcelo Blatt, Alexander Gusev, Yuriy Polyakov, Kurt Rohloff, and Vinod
  Vaikuntanathan.
\newblock Optimized homomorphic encryption solution for secure genome-wide
  association studies.
\newblock {\em BMC Medical Genomics}, 13(7):1--13, 2020.

\bibitem{LLK+20}
Yongwoo Lee, Joonwoo Lee, Young-Sik Kim, HyungChul Kang, and Jong-Seon No.
\newblock High-precision and low-complexity approximate homomorphic encryption
  by error variance minimization.
\newblock Cryptology ePrint Archive, Report 2020/1549, 2020.
\newblock \url{https://eprint.iacr.org/2020/1549}.

\bibitem{HighPrecisionFirst}
Joon-Woo Lee, Eunsang Lee, Yongwoo Lee, Young-Sik Kim, and Jong-Seon No.
\newblock Optimal minimax polynomial approximation of modular reduction for
  bootstrapping of approximate homomorphic encryption.
\newblock Cryptology ePrint Archive, Report 2020/552, Second Version, 2020.
\newblock \url{https://eprint.iacr.org/2020/552/20200803:084202}.

\bibitem{CKK+19}
Jung~Hee Cheon, Dongwoo Kim, Duhyeong Kim, Hun~Hee Lee, and Keewoo Lee.
\newblock Numerical method for comparison on homomorphically encrypted numbers.
\newblock In {\em ASIACRYPT 2019}, pages 415--445, 2019.

\bibitem{HE}
Kaiming He, Xiangyu Zhang, Shaoqing Ren, and Jian Sun.
\newblock Delving deep into rectifiers: Surpassing human-level performance on
  imagenet classification.
\newblock In {\em Proceedings of the IEEE International Conference on Computer
  Vision (ICCV)}, pages 1026--1034, 2015.

\bibitem{LLNK20}
Eunsang Lee, Joon-Woo Lee, Jong-Seon No, and Young-Sik Kim.
\newblock Minimax approximation of sign function by composite polynomial for
  homomorphic comparison.
\newblock Cryptology ePrint Archive, Report 2020/834, 2020.
\newblock \url{https://eprint.iacr.org/2020/834}.

\end{thebibliography}
}

\newpage
\appendix
\section{Appendix}
\subsection{Minimax Composition of ReLU}
Lee et al. \cite{MinimaxRelu} show that the ReLU function has to be approximated with sufficiently high precision if we use the pre-trained model parameters with the original ResNet-20 model. We need a polynomial with quite a large degree if the ReLU function is approximated by a single minimax polynomial, and it needs a pretty large running time to evaluate homomorphically. Instead of using the single minimax polynomial for the ReLU function, they use the formula $\mathrm{ReLU}(x) = \frac{1}{2}x(1 + \mathrm{sign}(x))$ and approximate $\mathrm{sign}(x)$ by the minimax composition of the small polynomials \cite{LLNK20}. It reduces the running time of the homomorphic evaluation of the ReLU function, and this approximation method makes more practical the homomorphic evaluation of the non-arithmetic functions such as the ReLU function.

\subsection{Analysis of Bootstrapping Failure}
We describe how the bootstrapping failure affects the evaluation of the whole ResNet, and propose a method to reduce the bootstrapping failure probability.
As the bootstrapping of the CKKS scheme is based on the sparsity of the secret key, there is a failure probability of its bootstrapping.

Here is the reason why approximated modulus reduction in the previous CKKS bootstrapping has a certain failure probability.
The decryption formula for a ciphertext $(a,b)$ of the CKKS scheme is given as $a \cdot s + b = m + e \pmod{\cR_q}$ for the secret key $s$; hence, $a \cdot s + b \approx m + q\cdot I \pmod{\cR_Q},$  where the Hamming weight of $s$ is $h$.
As the coefficients of $a$ and $b$ are in $[-\frac{q_0}{2}, \frac{q_0}{2}),$ we have that the coefficients of $a\cdot s + b$ have an absolute value less than $\frac{q_0(h+1)}{2}$.
However, by LWE assumption, the coefficients of $a\cdot s + b$ follow a scaled Irwin-Hall distribution and it is previously assumed that the coefficients of $I < K = O(\sqrt{h})$ \cite{LLK+20}.
As the modulus reduction function is approximated in the domain 
$\{0, \pm 1, \dots, \pm (K-1)\} \times (-\epsilon, \epsilon)$,
if a coefficient of $I$ has a value greater than or equal to $K$, the modulus reduction returns a useless value, and thus, failed.

$O(\sqrt{h})$ is a reasonable upper bound for a single bootstrapping, but it is not enough when the number of slots is large and there are many bootstrappings.
Let $p$ be the probability of modulus reduction failure, $\mathrm{Pr}(|I_i|\ge K)$.
If there are $n$ slots in the ciphertext, there are $2n$ coefficients to perform modulus reduction.
Hence, the failure probability of single bootstrapping is $1-(1-p)^{2n}\approx 2n\cdot p$.
Similarly, when there are $N_b$ bootstrappings in the evaluation of the whole network, the failure probability of the whole network is $2N_b \cdot n \cdot p$.
As there are many slots in our ciphertext, and thousands of bootstrapping are performed, the failure probability is very high when we use previous approximate polynomials.

\end{document}